\documentclass[letterpaper, 10 pt, conference]{ieeeconf}  

\IEEEoverridecommandlockouts                              

\usepackage{graphicx} 
\usepackage{subcaption}
\usepackage{epsfig} 
\usepackage{multirow}
\usepackage[
colorlinks=true,
urlcolor=blue,
linkcolor=black
]{hyperref}
\title{\LARGE \bf
	A Joint Motion Model for Human-Like Robot-Human Handover
}

\author{Robin Rasch$^{1}$, Sven Wachsmuth$^{2}$ and Matthias K\"{o}nig$^{1}$
	\thanks{*This work is financially supported by the German Federal Ministry of Education and Research (BMBF, Funding number: 03FH006PX5).}
	\thanks{$^{1}$Robin Rasch and Matthias K\"{o}nig are with Faculty Campus Minden,
		Bielefeld University of Applied Sciences, 32427 Minden, Germany
		{\tt\small firstname.surname@fh-bielefeld.de}}%
	\thanks{$^{2}$Sven Wachsmuth is with the Central Lab Facilities,
		Cognitive  Interaction  Technology  Excellence  Cluster,  Bielefeld  University,
		Germany. {\tt\small swachsmu@techfak.uni-bielefeld.de}}%
}

\begin{document}

\maketitle
\thispagestyle{empty}
\pagestyle{empty}

\begin{abstract}
 In future, robots will be present in everyday life. The development of these supporting robots is a challenge. A fundamental task for assistance robots is to pick up and hand over objects to humans. By interacting with users, soft factors such as predictability, safety and reliability become important factors for development. Previous works show that collaboration with robots is more acceptable when robots behave and move human-like. In this paper, we present a motion model based on the motion profiles of individual joints. These motion profiles are based on observations and measurements of joint movements in human-human handover. We implemented this joint motion model (JMM) on a humanoid and a non-humanoidal industrial robot to show the movements to subjects. Particular attention was paid to the recognizability and human similarity of the movements. The results show that people are able to recognize human-like movements and perceive the movements of the JMM as more human-like compared to a traditional model. Furthermore, it turns out that the differences between a linear joint space trajectory and JMM are more noticeable in an industrial robot than in a humanoid robot.
\end{abstract}

\section{INTRODUCTION}

Robots will be commonplace in the future and are designed to improve our quality of life at work and at home. Robots and humans will solve tasks together and interact with each other. These interactions can be handovers of an object between humans and robots. Handovers and other interactions can lead to direct physical contact between humans and robots. Therefore, it is important to take soft factors into account when developing robots that will operate in the personal space of humans. These soft factors are, for example, the traceability of the robot and the ease of use. Above all, the feeling of safety is important for interactions between humans and robots. For users, this feeling is restricted by the powerful movements of the robots and the associated potential danger. This requires that human safety, but also the feeling for it, must be taken into account in the development of robots. In order for an assistant robot to be able to give such a feeling and not cause any damage when interacting with people, it is necessary to create familiar situations for the user. This can be achieved by increasing the user's knowledge of the robot's movements. Two approaches can be used for this: training a model for each user or imitating human movements and behaviors.

As a starting point for our research, we assumed a basic scenario in which the robot fetches an object and delivers it to the user. Various research questions arise for this scenario: How does the robot know when to start handover, so that the user is able to react? In which pose should the robot place the object? How could the robot react to man's dynamic movements without colliding? When should the robot release the object so that it does not fall down or the user has to pull it? And how should the robot move so that the movement is safe and comfortable for the user? To sum it up, \textit{how} can a single task for an autonomous robot working in a person's personal space be developed, which gives the user a feeling of safety and takes comfort and predictability into account.

One aspect to increase the user's feeling of safety during interaction with robots is to make the movements known and predictable for the user. This can be made possible by transferring the movements of human-human interactions to robot-human interactions. The more the movements of a robot resemble those of a human being, the higher the feeling of safety~\cite{shibata95, huber08}. To achieve more robust and safer human-robot interactions, it is a goal to expand and discuss the knowledge about how humans interact with each other.

The movements during the transfer of objects have already been investigated~\cite{huber09, rasch17}. It turned out that people have different movement sequences for transferring objects depending on the situation and other parameters. Situations are determined by the combination of standing, sitting or moving interaction partners. One of the studies that dealt with this topic was Huber et al.~\cite{huber09}. A trajectory model for the wrist was developed from an experiment with two people sitting opposite each other. The model, which is approached in this paper, is based on the transfer of two standing interaction partners. The object is small and light and can be carried and transferred with one hand. This can be a small bottle or a pencil, for example. The study of our previous work~\cite{rasch17} served as the basis for the joint motion model (JMM). The results of the study are a trajectory model for the wrist, but also a general movement pattern of the other joints in the arm. However, a trajectory model has several disadvantages for implementation on a robot. On the one hand, the trajectory is very dependent on the kinematics or inverse kinematics of the robot. On the other hand, the number of degrees of freedom can lead to joint configurations that look very inhuman, for example an overstretched elbow joint. For this reason, our JMM used a model for each individual joint of the general movement sequence. An illustrative video can be found under the link\footnote{\url{https://youtu.be/oCd1sDV3PAs}}.


\section{RELATED WORK}
The question of how to make the handover of objects between robots and humans more human-like has already been investigated in various works. Our hypothesis that we could generate more human-like movements with a joint movement model was derived from these aspects: 1) the general approaches and processes of an object transfer, and 2) the movement models that exist for handover. The state of the art in these aspects is explained below.

\subsection{Handover Process}
Strabala et al.~\cite{strabala13} provide an overview of the temporal sequence of a transfer. This sequence can be divided into three steps. After the initial transport or carry phase, the intention phase follows. In this phase, different signals are sent out by one of the interaction partners, which trigger the actual transfer phase. These signals can be facial expressions, gestures or verbal communication.

The transport phase can vary in length. If the object is further away or if it is in another room, a path planner is also required. This planner must be aware of the human interaction partner to maintain the feeling of safety during this phase. Different approaches show how the object can be transported in a safe and user-friendly way~\cite{sisbot05,mainprice12}. The transport phase also affects the intention phase. An intention can be for example if a handing over person signals readiness by means of a gesture in which the person remains in a suitable position and extends the hands~\cite{lee11}. The gestures and facial expressions are also part of the next phase. Grigore et al.~\cite{grigore13} built a behavioral model for robots that takes into account the orientation of the head and the view of the interaction partner during a handover. Some of the basic research projects~\cite{lee11,basili09} of the transfer phase highlights a general process: 1) carrying, 2) coordinating and 3) object exchange. During the exchange phase, it is important for the robot to release the object at the moment of exchange. This moment is described by various factors. Some controllers focus on the facial expressions and gestures of the interaction partner~\cite{grigore13, Dragan13}, other controllers uses an approach to determine a stable grip. For this purpose, different sensor systems and algorithms or learning methods are used to determine this grasp in different works~\cite{kupcsik18, medina16, eguiluz17}. In the coordination phase, various parameters of handover are negotiated between the interaction partners, such as the handover position based on various criteria such as field of view, security and accessibility~\cite{Cakmak11, sisbot12}. Here the interaction is also determined by gestures and movements. Another factor of this interaction is the degree of synchronization. In contrast to the hand  shake~\cite{tagne16}, where synchronization determines the oscillating process, a handover requires an interpersonal synchrony~\cite{delaherche12}. Our approach deals with the carrying phase. In this position, the object is moved to the transfer position. In contrast to the transport phase, this movement is short and is usually carried out with the arms only.

\subsection{Motion Models}
Human-human studies have shown that the object is not moved along a linear trajectory during the carrying phase. Initial parts of a movement indicate the intention to hand over~\cite{strabala13}. The remaining movement sequence is used to position the object. The movement is examined for different scenarios and implemented with different models and methods. One of the first studies was Shibata et al.~\cite{shibata95}, which showed that a trajectory of a handover between two seated persons is not linear but follows a general pattern.

\begin{figure*}[h!]
	\centering
	\begin{subfigure}[b]{0.22\textwidth}
		\centering
		\includegraphics[height=3cm]{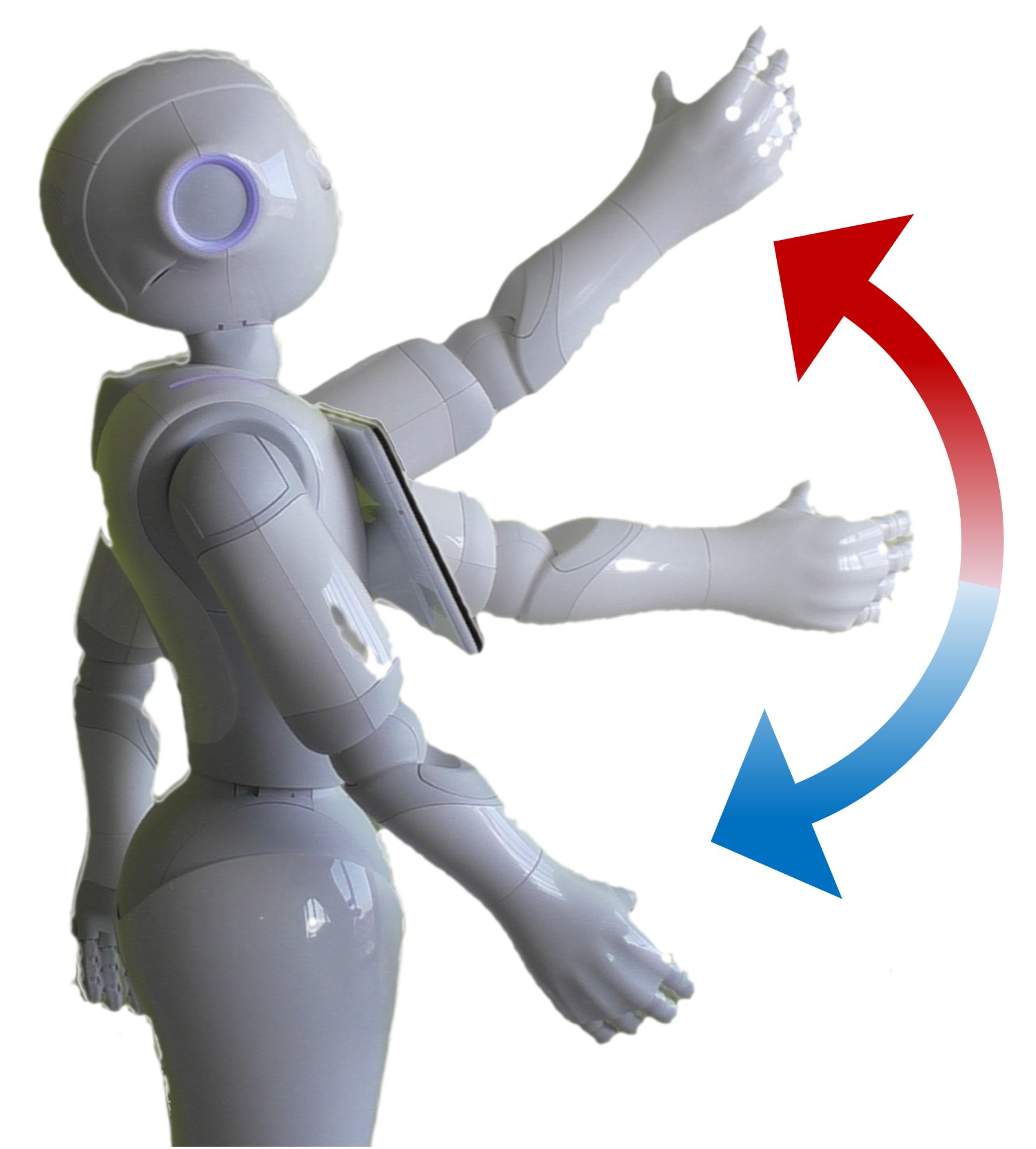}
		\caption{Shoulder flexion (red) and extension (blue)} 
		\label{fig:prim1}
	\end{subfigure}
	\hfill
	\begin{subfigure}[b]{0.22\textwidth}
		\centering
		\includegraphics[height=3cm]{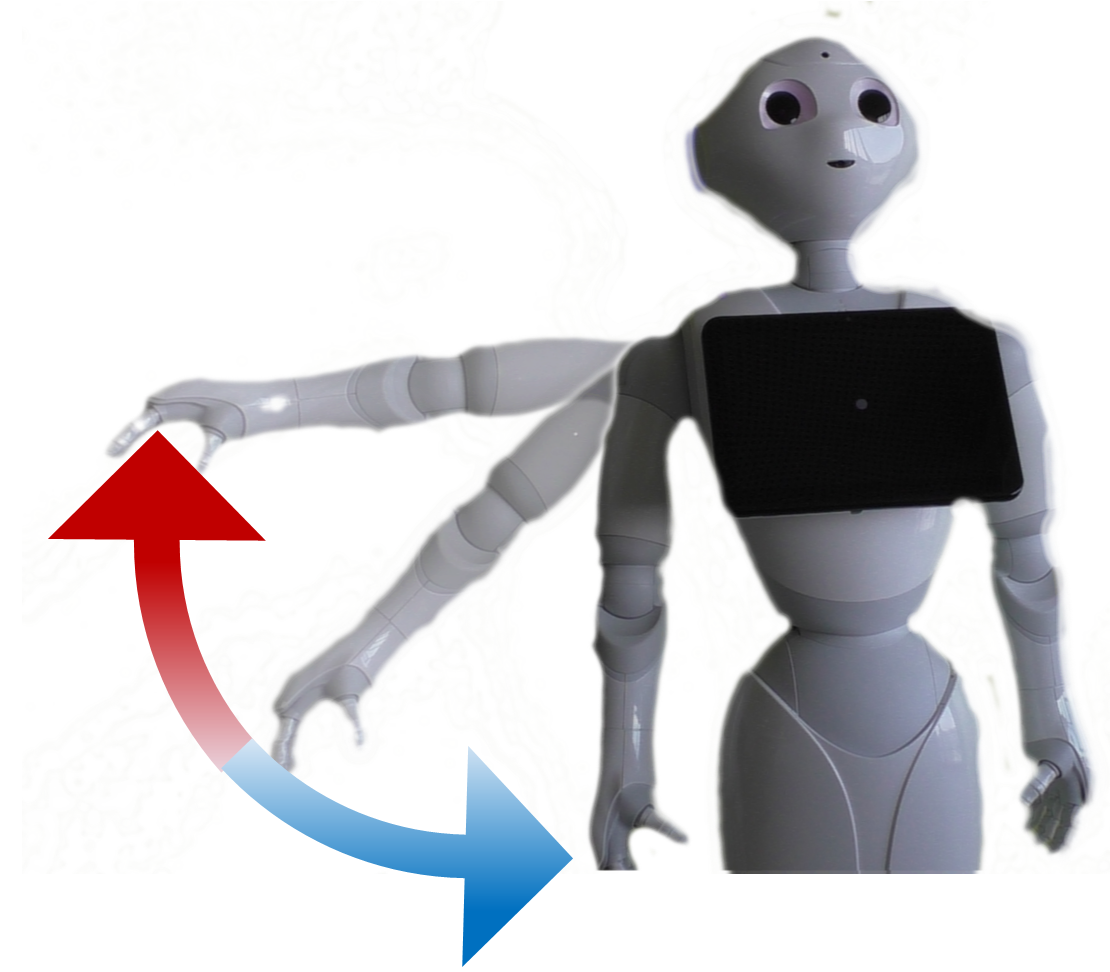}
		\caption{Shoulder abduction (red) and adduction (blue)}
		\label{fig:prim2}
	\end{subfigure}
	\hfill
	\begin{subfigure}[b]{0.22\textwidth}
		\centering
		\includegraphics[height=3cm]{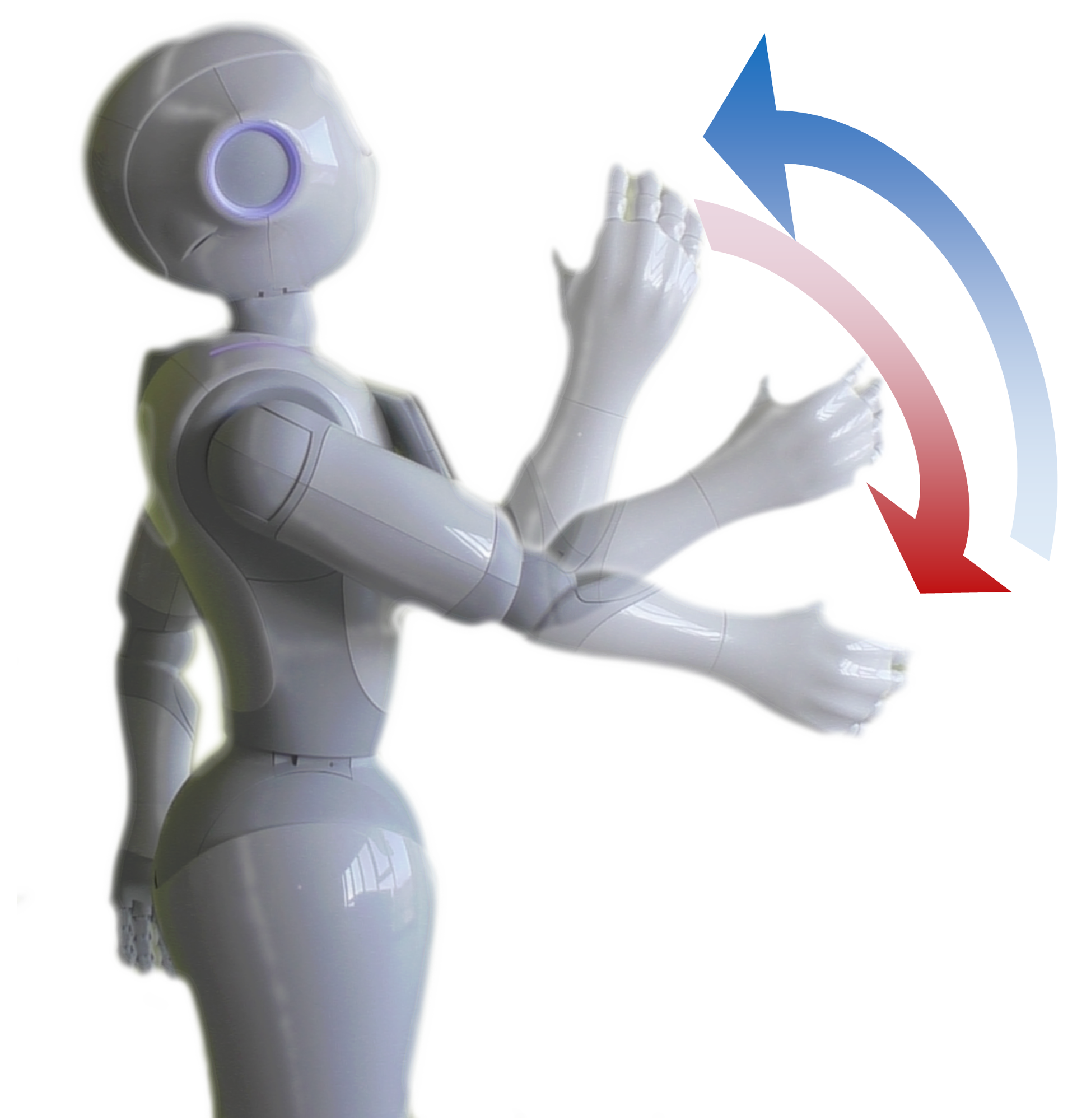}
		\caption{Elbow extension (red) and flexion (blue)}
		\label{fig:prim3} 
	\end{subfigure}
	\hfill
	\begin{subfigure}[b]{0.22\textwidth}
		\centering
		\includegraphics[height=3cm]{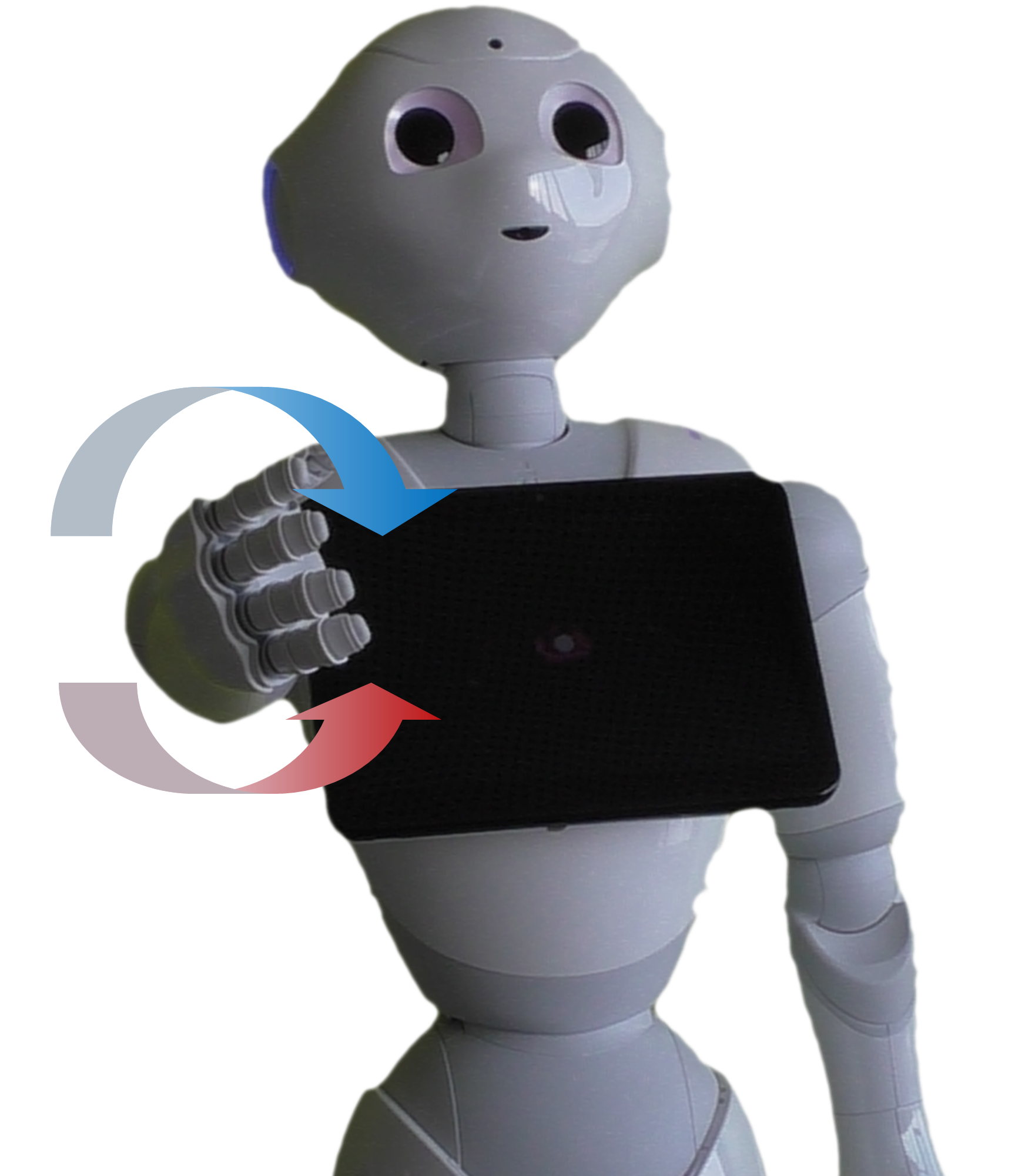}
		\caption{Forearm motions: supination (red) and pronation (blue)}
		\label{fig:prim4} 
	\end{subfigure}
	\caption{The motion primitives that are used during a transfer and their countermovements are executed by a Pepper robot. The combination of shoulder movements is also known as circumduction.}
	\label{fig:primitives}
\end{figure*}

An implemented planner is the trajectory planner from Huber et al.~\cite{huber09}, which is based on the minimum jerk approach and has been further developed to the decoupled minimum jerk trajectory generator. The validation with a human-robot study shows the advantages of such a planner. Our previous work~\cite{rasch17} was based on this planner. In contrast to Huber et al., the probants were not seated during the study, but stood opposite each other. The object was a light object that can be held in one hand. The results of this study were a trajectory model for the wrist and a general movement model consisting of movement primitives. Fig. \ref{fig:primitives} shows the motion primitives that are executed in parallel or sequentially. The total movement consists of a flexion of the shoulder (Fig. \ref{fig:prim1}), which together with an adduction (Fig. \ref{fig:prim2}) leads to a partial circumduction. A flexion of the elbow (Fig. \ref{fig:prim3}) extends the arm towards the receiver. Finally, there are two variations referring to the rotation of the forearm, shown in Fig. \ref{fig:prim4}.  In the first variant, the closed palm is turned downwards by pronation of the forearm. In variant two, the palm of the hand is turned upwards by a supination. Further short movements, e. g. an abduction of the shoulder at the beginning of the movement, can also be noticed during a handover. Based on this movement model, our paper presents a detailed movement model for the individual joints.

Different techniques have been used to create a movement model. One variant is to imitate the movement of people. For this purpose, the movement is recorded by various sensors and represented by mathematical or logical models. These models are optimized on the basis of the data according to defined criteria. Common sensor approaches have been camera-based approaches or magnetic field-based tracking. Inertial measurement unit sensors were used in our previous study. The criteria according to which a model is optimized are versatile, e. g. minimum-variance~\cite{Harris98} or minimum-jerk~\cite{Flash85}. The trajectory model in~\cite{rasch17} was based on a five-degree polynomial and was optimized after the minimal distance to the trajectories of human-human handover. 

\section{JOINT MOTION MODEL}
We considered the task of handing over an object to a human being with a robot. The robot should move as similarly as possible to humans to maintain the feeling of safety. We assumed that the object is light and small, so it could be carried and held with one hand. In addition, the object was rigid and the physical properties were known. The robot also had the object in hand and came from the transport phase. In our scenario, we did not consider coordination issues such as handover poses, stability, or signaling problems. 

We were interested in making the movement as similar as possible to human. In doing so, we did not focus on a human-like trajectory, as in previous works, but on a human-like movement pattern. The joint motion model was based on time-angle functions, whereby for each primitive movement of the handover a function was determined. To define and calculate the motion functions, the movements of people were analyzed during handovers with video data.

Furthermore, we were interested in the subjective perception of humans for the movements of the robot. Could a person perceive changes in the movements? Did the subject notice a human likeness? Did anything change in the statements when you point out differences? To answer these questions, we implemented the motion model on two robots and interviewed test persons.

\subsection{Motion Analysis}
\begin{figure}[h!]
	\centering
	\includegraphics[width=0.8\columnwidth]{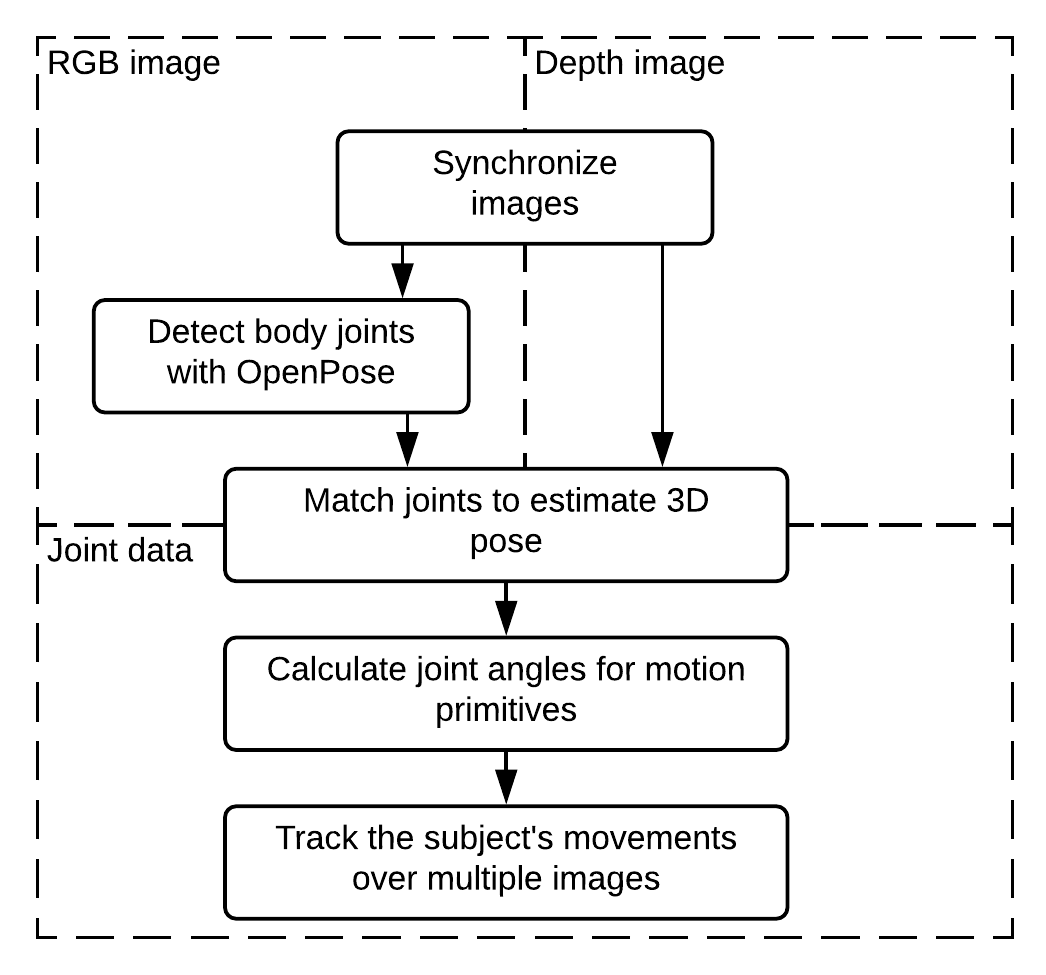}
	\caption{Analysis process for joint angles from RGB-D video data.}
	\label{fig:analysis}
\end{figure}

\begin{figure}[h!]
	\centering
	\includegraphics[width=\columnwidth]{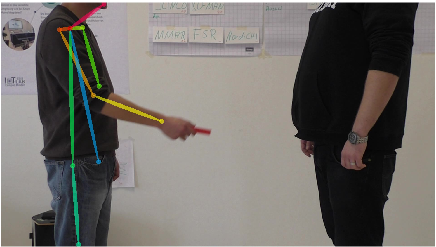}
	\caption{Sample image of the video analysis with the RBG camera. The left person hands over an object with his right hand. The positions of the joints and limbs have been detected and highlighted with OpenPose.}
	\label{fig:vid_demo}
\end{figure}

The video data were created on the basis of two experiments. The first dataset was created from the human-human study from a previous study 
\cite{rasch17}. Here, 150 handovers were carried out by 26 people. The subjects were aged between 16 and 49 years. The camera was positioned to the right behind the giving subject. In contrast to the previous study, another video process was used to analyze the data. For this we used a technology that was already used by Sprute et al.~\cite{sprute18} for the analysis of gestures. Our analysis process is based on OpenPose and Convolutional Pose Machines (CPM)~\cite{cao17} and is depicted in Fig.~\ref{fig:analysis}. First, our implementation for ROS synchronized RGB and depth images. CPM then detected the keypoints on the RGB image and estimated the body joints, as shown in Fig.~\ref{fig:vid_demo}. The points were matched with the depth image to determine the 3D pose of the joints. The angles of the movement primitives were then calculated and recorded. Since OpenPose does a single-frame computation, an additional tracking algorithm had been implemented to track the right person on the image.

Since the RGB-D camera only had a frame rate of 30 frames per second, a second data set was recorded. A simple RGB camera with a shooting rate of 100 FPS and a resolution of 1920x1080 pixels was used. Since it did not collect 3D data, the analysis process and scenario settings had been adjusted. At first, the angle of view was set perpendicular to the transfer. In this way, further information on the flexion/extension of the shoulder and elbow could be collected. Data on shoulder abduction or adduction could not be acquired from this perspective. The lack of depth information also eliminated the steps of synchronization and matching of the process chain. The tracking process remained, as both test subjects were still in the picture. The advantage of this recording method was the higher resolution, both in terms of image and time. The recorded second data set was considerably smaller than the first one. It included 42 handovers of seven test persons aged between 25 and 33 years. 

To determine a suitable movement model for the joints, several variable factors had to be deducted. The total time and thus the speed of the movements was standardized for the analysis. Since the persons determined the end positions themselves during the handover and human kinematics differed from person to person, the angles were considered both normalized and non-standardized. The data were also smoothed with a Savitzky-Golay filter to compensate for the discrepancies in CPM estimates.

The analysis of the data showed different motion profiles for the different joints. Since the different movements for the forearm were already examined in more detail in the previous work, the analysis of the elbow and shoulder is described in more detail in the following. First, the elbow flexion and extension is discussed.

\begin{figure}[h!]
	\centering
	\begin{subfigure}[b]{0.7\columnwidth}
		\centering
		\includegraphics[width=\columnwidth]{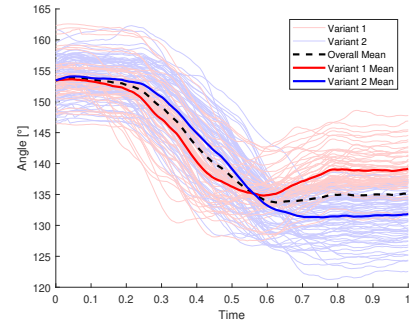}
		\caption{Only-time standardized elbow angles.} 
		\label{fig:elbow_angle_1}
	\end{subfigure}
	\hfill
	\begin{subfigure}[b]{0.7\columnwidth}
		\centering
		\includegraphics[width=\columnwidth]{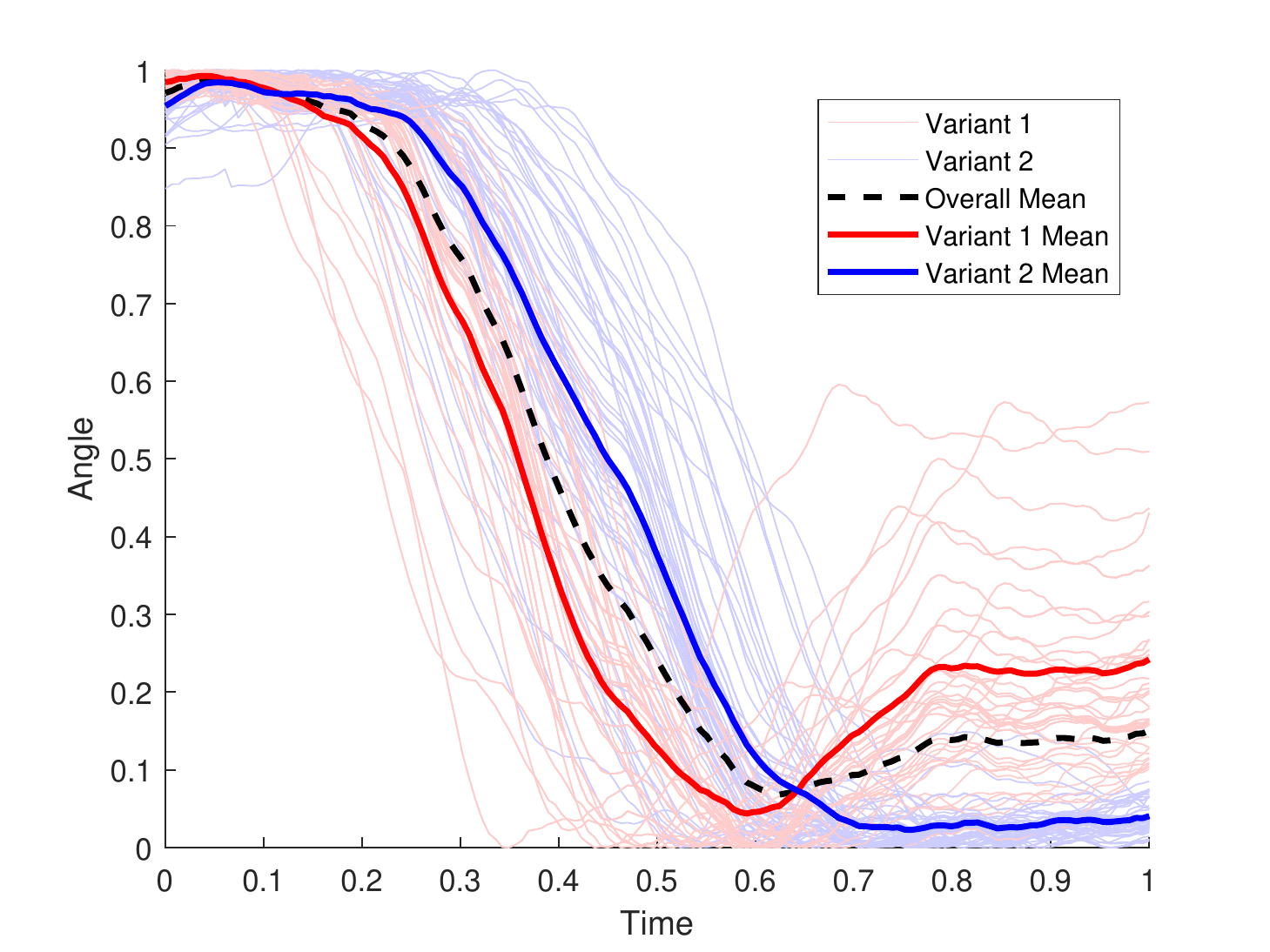}
		\caption{Time and ampiluted standardized elbow angles.}
		\label{fig:elbow_angle_2}
	\end{subfigure}
	
	\caption{The totality of all recorded elbow angles movement profiles. The two variants are marked by one color (blue and red). In addition, the overall average (black) and the averages for both variants (bold) are shown.}
	\label{fig:elbow_angle}
\end{figure}

The analysis showed two different variations of the motion sequence, which are illustrated and differentiated in Fig. \ref{fig:elbow_angle}. One variant $V_1$ showed a pronounced sub-motion, which was weak or non-existent in the other variant $V_2$. The basic procedure was the same for both variants. The angle was reduced by a flexion of the elbow. This raised the hand or end effector to a higher position. This mean that the shoulder joint did not cover the entire height range. In $V_1$, shown in the illustrations in red, the flexion was more pronounced at the beginning. The transferor drew the object closer to his body. To compensate this movement, an additional extension of the elbow followed at the end. This caused the object to move in the direction of the receiver. This additional movement is clearly visible in Fig.~\ref{fig:elbow_angle_1} by the global minimum of the movement to the middle of the execution. While $V_2$ generally descended monotonously, $V_1$ did not show this property. The evaluation also showed that the movements of $V_1$ end up with a higher, i. e. stretched out, arm. This becomes visible in the standardized view in Fig.~\ref{fig:elbow_angle_2}. Here it also becomes obvious that the movements of $V_1$ more scattered in their relative end positions. The absolute end-position showed similar behavior, as shown by the mean values in Fig. \ref{fig:elbow_angle_1}. A further characteristic was that the flexion in $V_2$ reaches the larger angle (avg. $V_1 : 18^\circ$, $V_2 : 22^\circ$), while in $V_1$ the minimum angle was reached relatively earlier. The distribution between the two variants was approximately the same. This could not be determined exactly, since the extension occurs to varying degrees. As a result, the transition between the two variants could not be determined unambiguously.

\begin{figure}[h!]
	\centering
	\includegraphics[width=0.7\columnwidth]{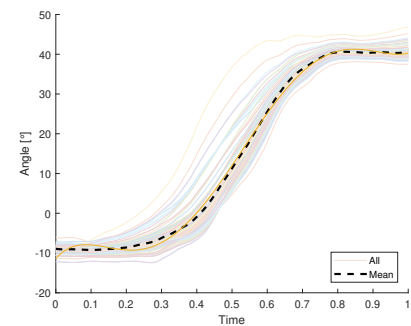}
	\caption{All recorded shoulder flexion and their mean value.}
	\label{fig:shoulder_angle}
\end{figure}

In contrast to the movement of the elbow joint, the movements of the shoulder joints are very uniform, see Fig.~\ref{fig:shoulder_angle}. The flexion of the shoulder starts slowly before reaching the maximum speed. Towards the end, the movement decelerates slowly. The amplitude of the movement is between 45 and 55 degrees. The monotonous gradient shows that the movement consists only of a flexion and not an extension. This movement leads to the elevation of the object into the end position, whereby the position height also depends on the angle of the elbow. 

The analysis of the adduction and abduction could only be carried out with images from the depth camera. It turned out that the profile was similar to flexion. The performed adduction guided the object in the horizontal orientation, whereby the transmitter moved the object in the direction of its own centre of the body and thus in the direction of the receiver. For both shoulder movements, their end positions depended on the end position of the handover. 

\subsection{Motion Model}
To transfer the results of the analysis into a model for robots, our approach determined a motion function for each primitive movement. The characteristic of the function was based on the analysis of the data. For this purpose, a  functions were assumed and the coefficients for the individual primitive motions were calculated by approximating the mean values of the analysis. By normalizing the data in terms of amplitude and time, the motion functions can be used for different speeds and start-, as well as end poses. But since the models also depends on the kinematics of the robot, additional constants were added. To adapt the models to the data as accurately as possible, a standard mathematical function was selected for each motion primitive, which matches the characteristics of the analysis data.

Due to the shape of a gooseneck, a sigmoid function~(\ref{eq:sig}) was chosen for the movement function of the shoulder. Compared to a polynomial function, it has the advantage to run at the edges against the constraints. 
\begin{equation}
\label{eq:sig}
f(x) = \frac{a}{b+e^{(-c*x)}}
\end{equation}
The coefficients a = 0.000905, b =  0.0008908 and c = 12.87 were determined by fitting the standardized data against time using the Levenberg-Marquardt algorithm. A sum of squared errors of $SSE = 0.0167$ and a coefficient of determination of $R^2 = 0.9994$ were achieved. With the additional parameters and robot constraints, the movement model for the joint angle $J_S$ resulted:
\begin{equation}
\label{eq:js}
J_S(t) = \frac{a * (j_e - j_0) * r_c}{b+e^{\frac{-c*t}{t_e}}}  + j_0,
\end{equation}
where $j_0$ denotes the start angle of the joint at start and $j_e$ the end angle of the joint. $r_c$ describes a damping factor, which is restricted by the robot kinematics and adjusted to it. The time $t$, or rather the speed, is expressed by the term ~$\frac{t}{t_e}$ and can be viewed relatively by looking at the end time~$t_e$.

For the motion function of elbow flexion, a seven-degree polynomial function (\ref{eq:poly}) was selected to cover the characteristics of both variants. 

\begin{equation}
\label{eq:poly}
f(x) =  c_7x^7 + c_6x^6 + \ldots + c_1x + c
\end{equation}

Instead of determining the coefficients of the mean value, the coefficients $c, c_1, \ldots, c_7$ for the mean values of the two variants were determined.  To determine the mean values, the recorded data were divided into the two variants. The time position of the global minimum and the difference to the end of the movement were considered as criteria. By using a polynomial to map the analysis data, the resulting function is only valid within the limits~$[0,1]$. The coefficients for the variants are shown in Table \ref{tab:coef}. Variant one achieves $SSE = 0.0197$ and $R^2 = 0.9989$. Variant two has $SSE = 0.0112$ and $R^2 = 0.9994$ with respect to the analysis data.
\begin{table}
	\centering
	\footnotesize 
	\begin{tabular}{|c|c|c|c|c|c|c|c|c|c|}
		\hline
		Var. & \bfseries $c_7$ & \bfseries $c_6$ & \bfseries $c_5$ & \bfseries $c_4$ & \bfseries $c_3$ & \bfseries $c_2$ & \bfseries $c_1$\\
		\hline\hline
		1 &  23.2 & 34.2 & -240.9 & 314.6 & -157.3 & 27.2 & -1.7 \\
		2 & 115.1 &	-376.9  & 454.1  & -240.1 & 53.1 & -6.7  & 0.5\\
		\hline
	\end{tabular}
	\caption{Polynomial coefficients for both variants of elbow motions. $c = 1.0$ for both variants.}
	\label{tab:coef}
\end{table}
Using the previous constants, parameters and limitations, this resulted in the function for the movement model of the elbow joint:
\begin{equation}
\label{eq:je}
J_E(t) = j_0 + (j_0 - j_e)r_c(c_7(\frac{t}{t_e})^7 + c_6(\frac{t}{t_e})^6 + \ldots + c),
\end{equation}
where $r_c, j_0, j_e$ have the same meaning but are not identical to~(\ref{eq:js}). For the further primitive movements, the same approaches were used to determine the motion functions. A sigmoid function was also used as the basis for adduction, while a polynomial was used for pronation or supination. 

\subsection{Implementation of Model}

\begin{figure}[h!]
	\centering
	\begin{subfigure}[b]{0.47\columnwidth}
		\centering
		\includegraphics[width=\columnwidth]{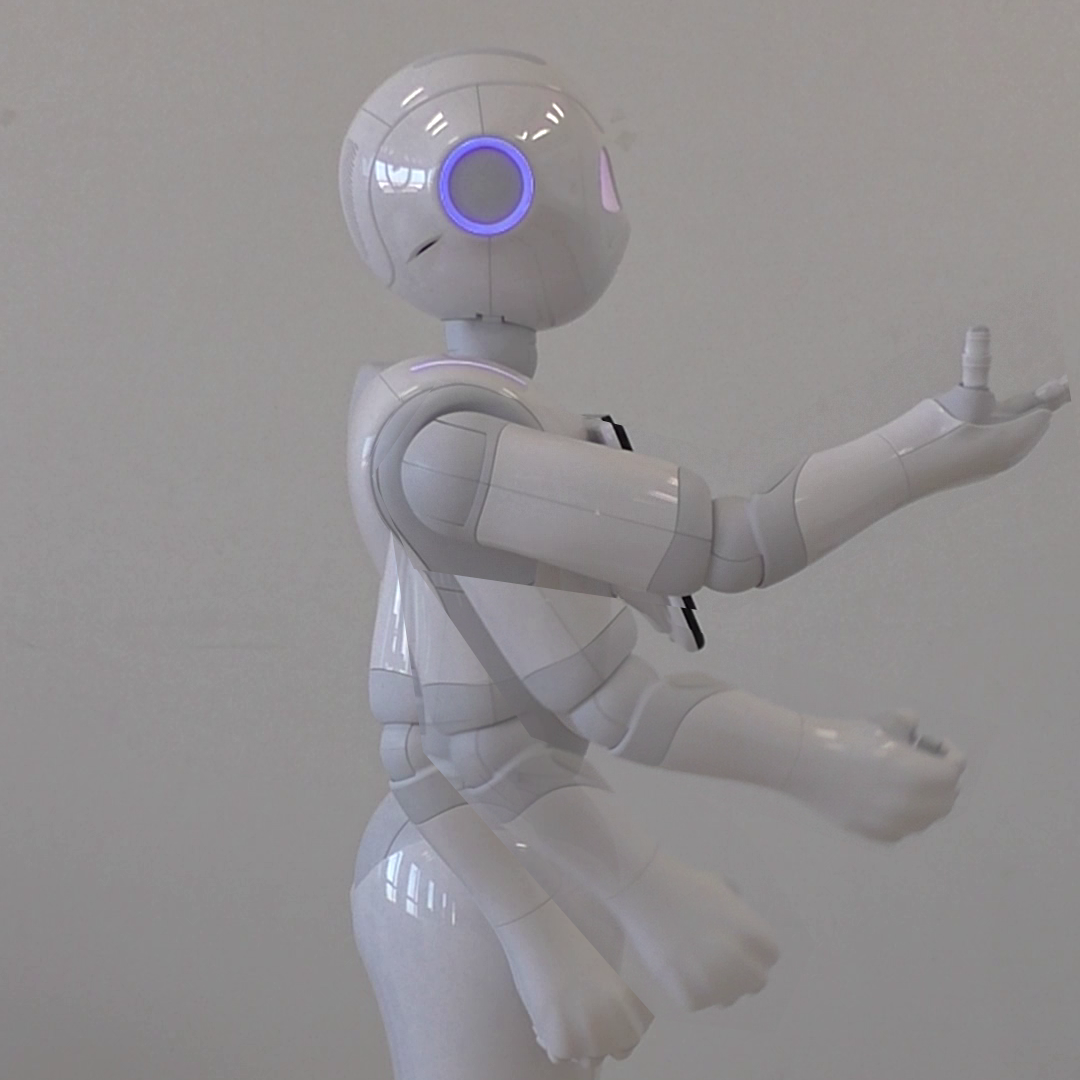}
		\caption{Linear joint space trajectory} 
		\label{fig:pep_dir}
	\end{subfigure}
	\hfill
	\begin{subfigure}[b]{0.47\columnwidth}
		\centering
		\includegraphics[width=\columnwidth]{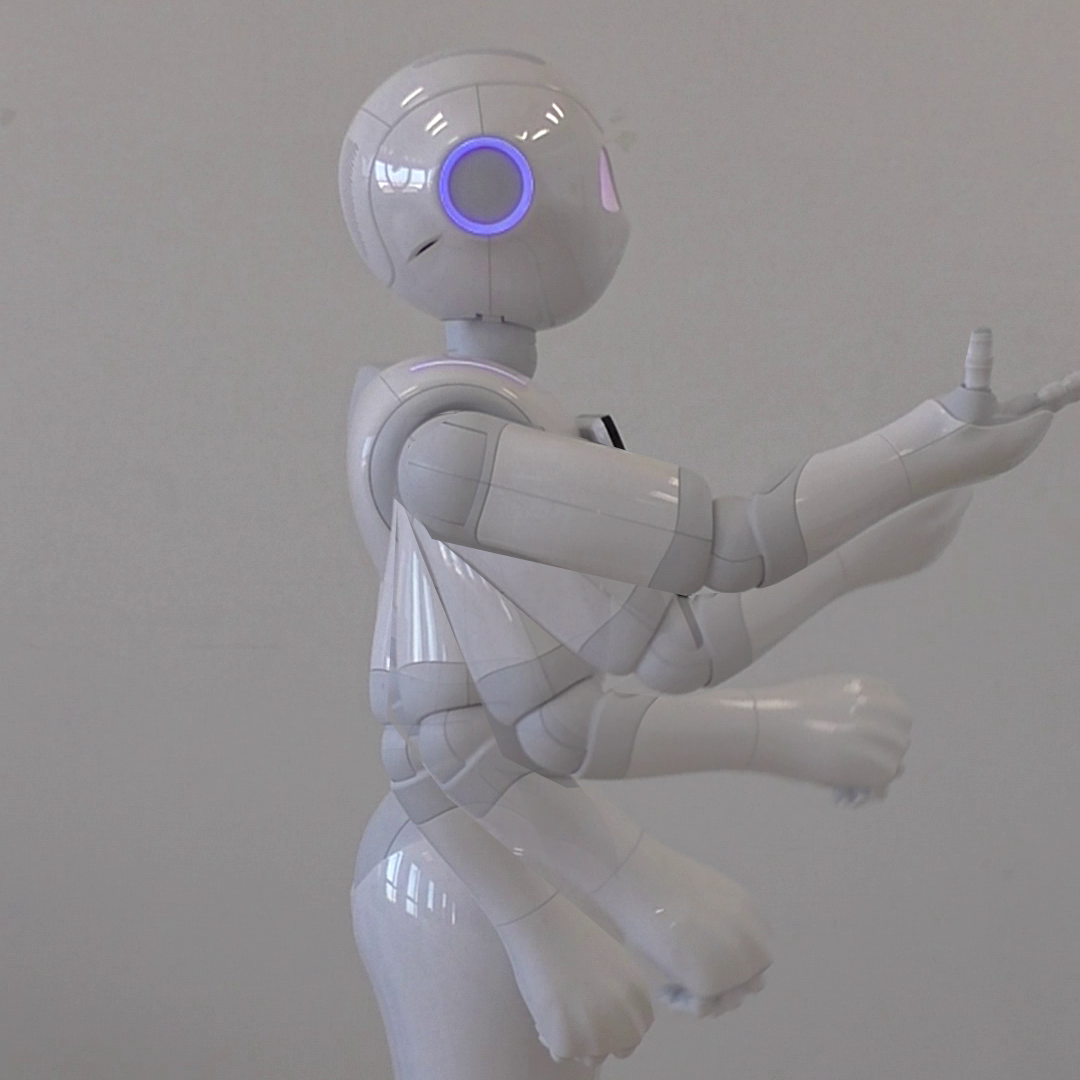}
		\caption{Joint motion model }
		\label{fig:pep_jmm}
	\end{subfigure}
	
	\begin{subfigure}[b]{0.47\columnwidth}
		\centering
		\includegraphics[width=\columnwidth]{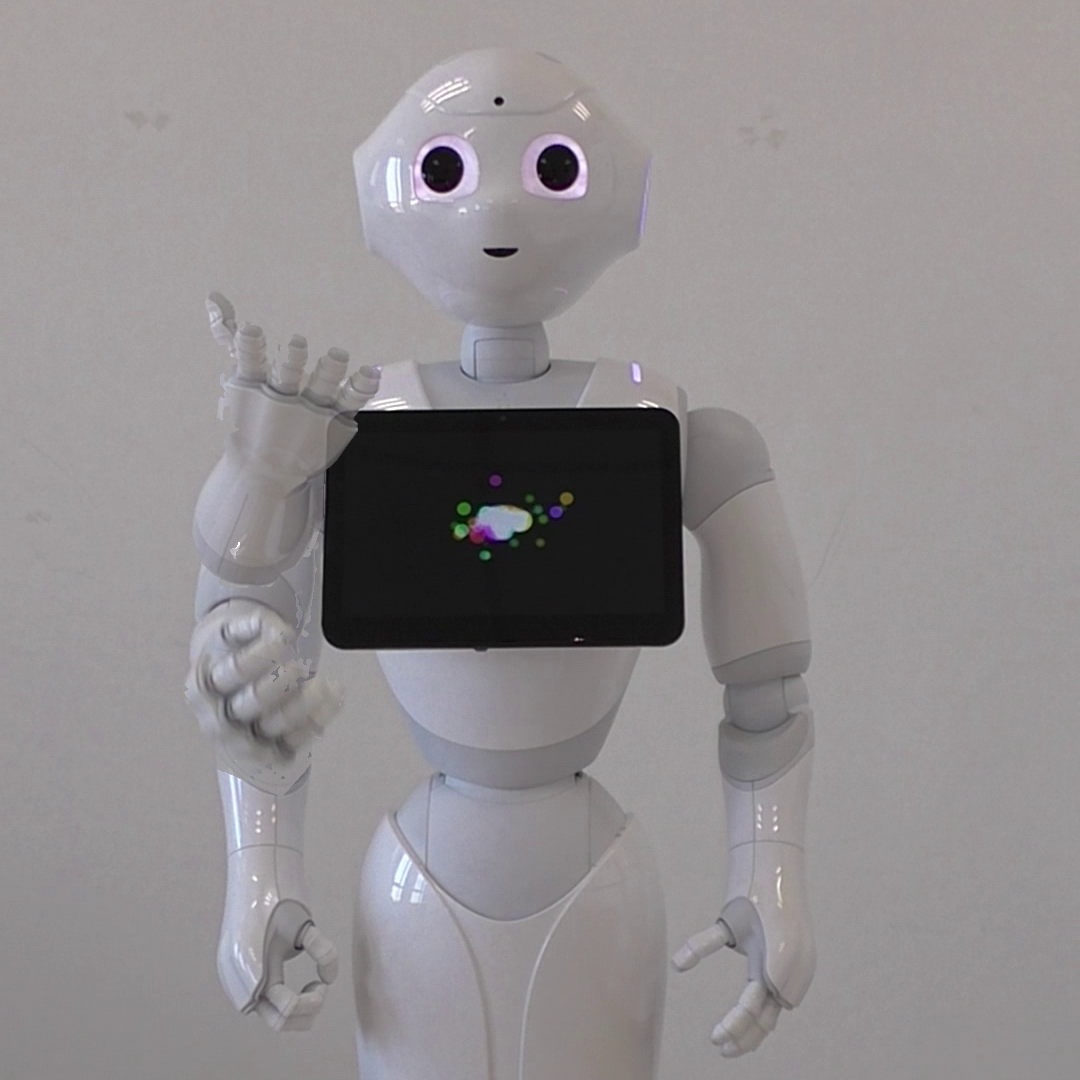}
		\caption{Linear joint space trajectory} 
		\label{fig:pep_dir_front}
	\end{subfigure}
	\hfill
	\begin{subfigure}[b]{0.47\columnwidth}
		\centering
		\includegraphics[width=\columnwidth]{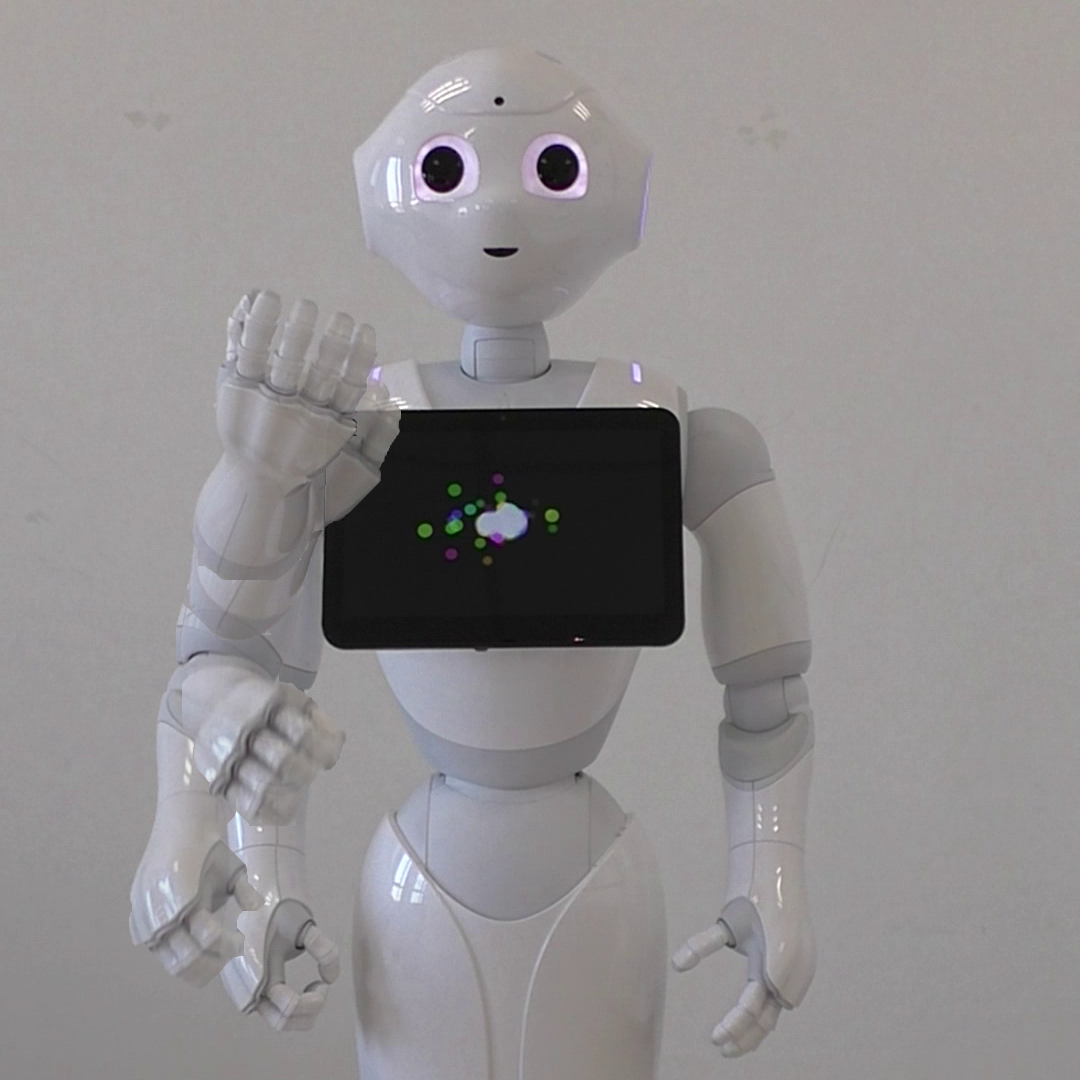}
		\caption{Joint motion model}
		\label{fig:pep_jmm_front}
	\end{subfigure}
	
	\caption{Image series of the two motion models executed by Pepper from different views.}
	\label{fig:pepper_motion_models}
\end{figure}

For the implementation of this model, it was necessary to control the joints as well as mapping the robot joints to the primitive movements. This is easy for humanoid robots as long as there are enough degrees of freedom. In our example this was done with a Pepper robot. In this case, the movements could be assigned directly to the joints. Only the pronation/supination of the forearm was represented by a rotation of the wrist. The robot constants $r_c$ were configured by start and end positions of handover. Based on kinematic properties (ratio between lower arm and upper arm) or limitations (tablet computer in front of the chest), the constant was used as stretching or damping factor. The different movements of the Pepper robot are shown in Fig. \ref{fig:pepper_motion_models}. The differences in movement are visible in the deflection of the elbow joint in the side view and the trajectory of the hand in the front view.

\begin{figure}[h!]
	\centering
	\begin{subfigure}[b]{0.4\columnwidth}
		\centering
		\includegraphics[width=\columnwidth]{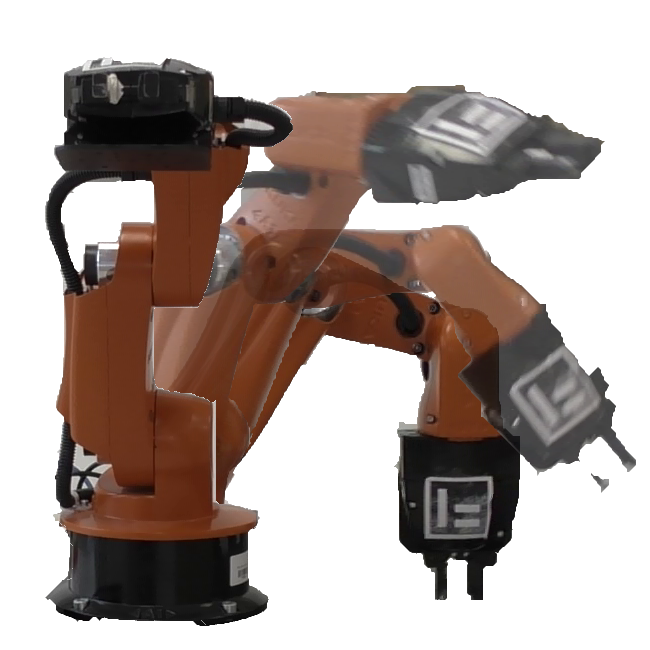}
		\caption{Linear joint space trajectory} 
		\label{fig:yb_dir}
	\end{subfigure}
	\hfill
	\begin{subfigure}[b]{0.4\columnwidth}
		\centering
		\includegraphics[width=\columnwidth]{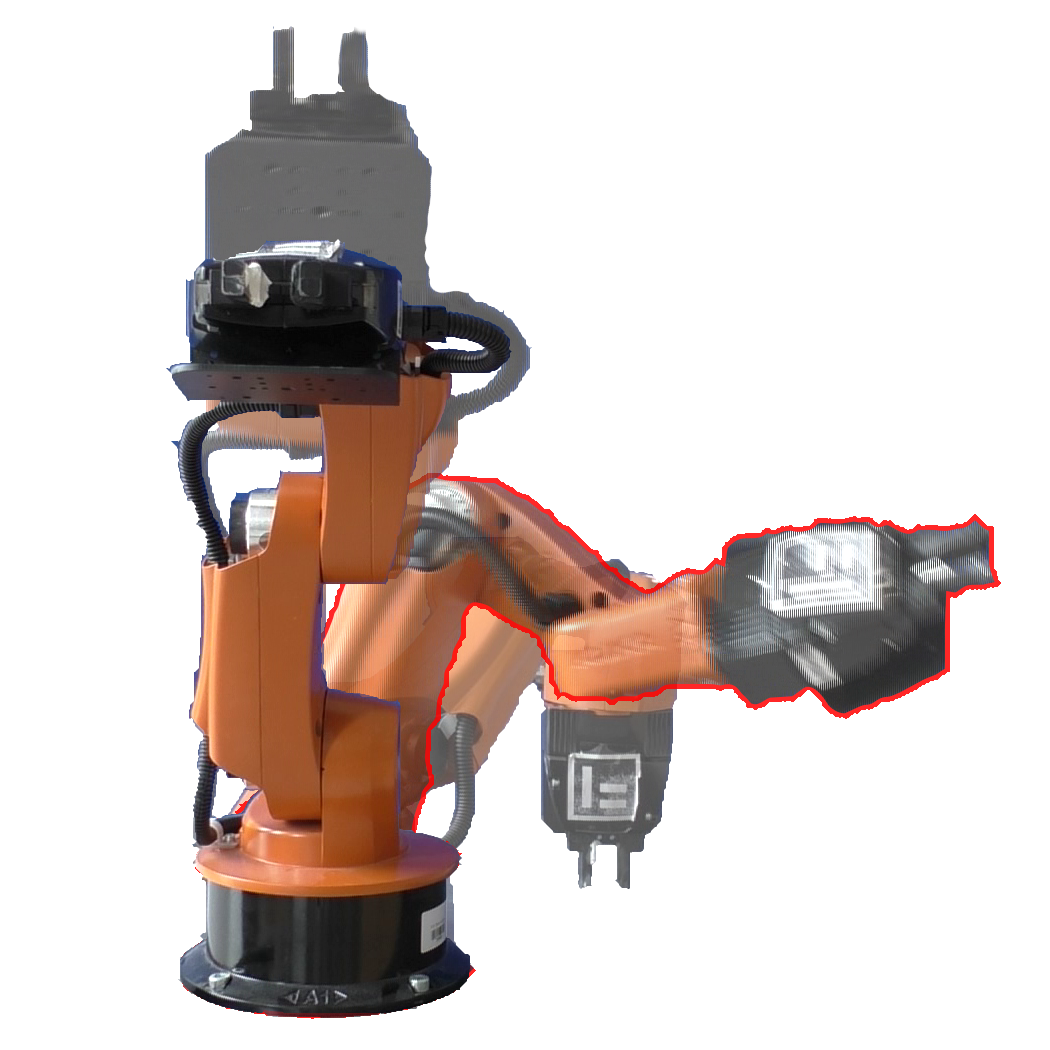}
		\caption{Joint motion model }
		\label{fig:yb_jmm}
	\end{subfigure}
	
	\caption{Image series of the two motion models executed by the industrial manipulator. The red marking shows the additional extension of the elbow at the JMM, which is clearly recognizable.}
	\label{fig:yb_motion_models}
\end{figure}

In our second application case, the motion pattern was implemented on an under-actuated robot manipulator, a Kuka Youbot. In this case, the mapping of the primitive movements to the joints was more complex, because the robot has only five degrees of freedom and the joints are not arranged in a human-like way. Since the first and last joint of the manipulator are twisting joints, they were mapped to the two twisting movement primitives. Joint 1 thus corresponded to the shoulder adduction and joint 5 to the pronation and supination of the forearm. The two flexions were represented by the rotating joints of the robot. Joint 3 represented the flexion of the shoulder and joint 4 the flexion and extension of the elbow. Fig. \ref{fig:yb_motion_models} shows the manipulator with both motion models. The configuration of the robot constant was carried out similarly to that of the humanoid robot.

\section{EXPERIMENTAL EVALUATION}
Initially, the question was raised as to whether an uninvolved and uninformed person could perceive changes during the movement of a handover and whether he or she could recognize a similarity to humans in the movement. Consequently, we conducted an experimental study to answer these questions.

\subsection{Experimental Setup and Procedure}
Twenty-five people aged between 22 and 34 participated in the study. The model was implemented on two robots. The Pepper robot served as a humanoid variant. The Kuka Youbot manipulator was used as a non-humanoid variant. This was placed on a 70 cm high table to reach a practical overall height. The test person was placed 50 cm in front of the robot, but was allowed to move freely in order to change the perspective on the robot.  The robots had a light object in the gripper before each sequence and were in a typical pose to start the transfer. 

The participants were divided into two groups without their knowledge. The experimental group consisted of 15 persons, the control group of 10 persons. The assignment was random. In both groups, the experiment was divided into two phases.

\paragraph{Phase 1}
The subjects were not informed about the subject and question of the study. They were instructed about the procedure of the experiment and that the robot wanted to hand over an object. The robots then demonstrated their movements. The sequence of the robots and the sequence of the movement model was randomly selected per person to exclude temporal relations and preferences. A linear joint space trajectory model model (LJST), was used as a counter model to our movement model \cite{corke17}. For the elbow flexion of the JMM we have decided on the variant 1, with the additional extension, to make the difference between the movements more visible. In the experimental group, each movement model was executed twice in succession. Between the two models there was no change between the robots. The control group, in which only the LJST was shown four times, is different.  After all robots and movements were demonstrated, the participants were asked to complete a questionnaire. After completing the questionnaire, the second phase followed.

\paragraph{Phase 2}
The subjects were instructed more precisely for this phase. Every subject was told the same thing. Translated from German: \textit{"Note the differences in the movements of the arm and joints, for example the shoulder or the elbow. See what looks more human-like to you."}  The exact differences between the movements were \textit{not} explained to them. Nor were any statements made about the similarity to humans of the models. In this phase, the movements of both robots were also demonstrated. The experimental group saw the two different movement models, while the control group evaluated the same movement twice for each robot. The same questionnaire as in phase~1 was answered again with an additional question about safety. 

In this questionnaire, the participants should answer various aspects of the research question subjectively. The following questions have been translated from German:
\begin{enumerate}
	\item \textit{Difference:} Have you noticed any difference in the movements of the humanoid/ industrial robot?
	\item \textit{Human Likeness:} Which movement of the humanoid/ industrial robot was most similar to humans?
	\item \textit{Detail:} How did you identify the likeness to humans?
	\item \textit{Robot comparison:} Which robot moved more like humans?
	\item \textit{Safety:} How safe did you feel about movement 1-4?
\end{enumerate}

The subjects were able to quantify the differences and safety using a Likert scale. One of the robots could be selected by the subjects in the human-like manner. The \textit{detail} question could be answered with a free text. Finally, the data on age, size and gender were collected.

The separation into control and experimental groups was undertaken to rule out the possibility that the explanation between phase 1 and phase 2 might have too much influence on the subjects and that they only follow the instructor's explanations.

\subsection{Results and Discussion}

%

\begin{figure}[h!]
	\centering
	\includegraphics[width=0.95\columnwidth]{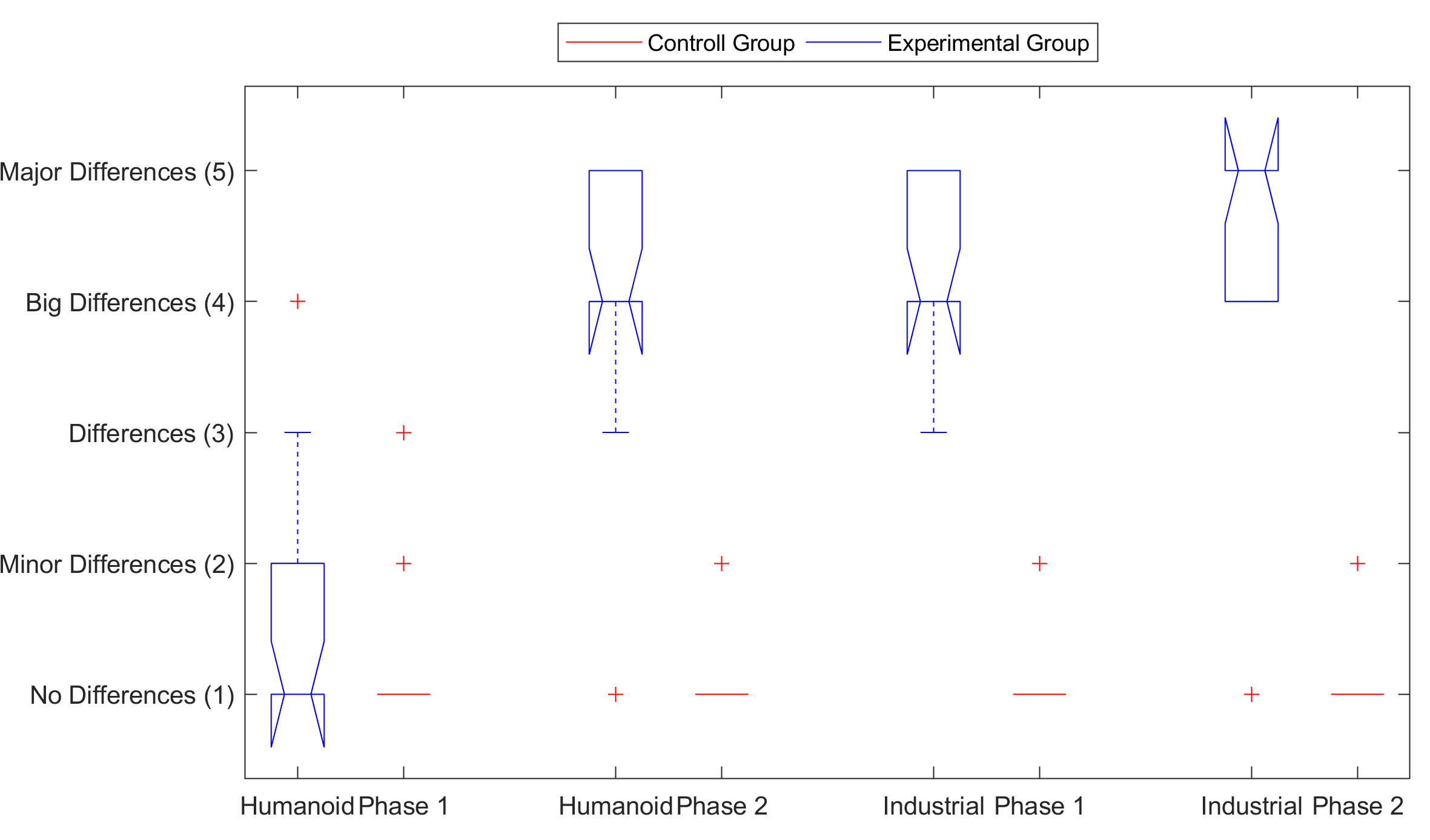}
	\caption{The development of the perceived differences between the first and second phase of the humanoid robot (HR) and the industrial manipulator (IR).}
	\label{fig:resbox}
\end{figure}

The results for the single phases and the different robots and movement models are shown in Fig. \ref{fig:resbox}. In phase one, almost all the participants in the experimental group find no (66.6\%) or just minor (13.3\%) differences in the movements of the humanoid robot. In contrast to the humanoid robot with an average value of 1.6, the test persons recognize a difference in the industrial manipulator with an average value of 4.3. This shows that a difference in the motion model is not perceived by an uninvolved person on a humanoid robot. This could presumably be attributed to the small differences in movement. The difference became clearer with the industrial robot, as the direct movement of the joints made a big difference to the motion model due to the kinematics. 73\% of subjects did not make any statements about the human-likeness of the movement of the humanoid robot, as they did not perceive any difference between the movements. The other part felt the JMM (75\%) more human-like than the LJST (25\%). This was different with the industrial robots. There all test persons gave an answer. 73.3\% felt the JMM more human-like and 26.6\% the LJST. The control group showed the expected results in comparison. 90\% of subjects saw no to little differences in the humanoid robot and 100\% in the industrial robot. Only one person saw differences and felt the LJST of the industrial robot and our model of the humanoid robot as human-like. The final question for phase one, which robot moves more humanoidly, was answered by 96\% of all subjects with the humanoid robot and 4\% with the industrial robot.

After the second phase, the test persons answered the questions again. The development of the perceived differences of the experimental group was decisive for our first hypothesis. After the explanation, 13.3\% of the subjects noticed differences, 46.6\% several differences and 33.3\% strong differences in the movement of the humanoid robot. For the industrial robot, 40\% noticed several differences and 53.3\% strong differences in movements.

To statistically verify the differences in the perception between the two phases, a paired-samples t-test was conducted for each robot in phase 1 and phase 2. For the humanoid robot there was a significant difference in the scores for phase~1~($M=1.6, SD=0.98$) and phase~2~($M=4.0, SD=1.06$); $t(14)=-7.159; p=0.00$. For the industrial robot there was no significant differences in the scores between phase~1~($M=4.28, SD=0.61$) and phase~2~($M=4.28, SD=1.06$); $t(14)=0.0, p=1.0$. The results of the control group had to be compared in order to check whether the trend was only based on the explanation of the instructor or on the perception of the subjects. A comparison of arithmetic averages and standard deviations showed that the trend did not occur in the control group. With the humanoid robot, the evaluation dropped from 1.3 (SD: 0.48) to 1.1 (SD: 0:18) and with the industrial robot, the average value remained at 1.1 (SD: 0.18). It could be deduced from this that subjects were able to perceive differences in the movements of robots. If the differences were too small, they could only be noticed when the subjects took a closer look at the movement.

After the differences in movements were recognized, the subjects also made more statements about the human similarity of the movement models. After phase two, 93\% of the subjects in the experimental group made a statement on the human resemblance of the humanoid robot and 100\% of the industrial robot. In the humanoid robot, 86.6\% of the subjects perceived the JMM as more humanoid and 6.6\% the LJST. The values of the industrial manipulator changed slightly compared to phase 1. 33.3\% found the LJST and 66\% the JMM to be more human-like. A binominal test indicated that the subjects perceived our proposed model significant more human-like if you do not consider the robot type ($p = 0.001$, 1-sided). It followed that our proposed model was perceived as  more human-like, both for the industrial robot and for the humanoid robot. When asked why a movement was perceived as more human-like, seven subjects answered that the sequence was more similar to humans, five gave the elongation of the arm at the end of the movement as the reason, four subjects found the reference model to be unnatural. Further answers related to speed and certain joint positions.
In addition, the subjects stated that they had felt the movements of the humanoid robot to be more human-like (93.3\%) than those of the industrial robot (6.6\%).

A paired-samples t-test was conducted to compare the feeling of safety in conditions of using JMM and LJST. The data of the experimental group were used. For both robots together there was a significant difference in the scores for JMM ($M=4.1, SD=0.88$) and LJST ($M=3.8, SD=0.96$); $t(29)=2.757, p = 0.01$. These results suggest that the JMM really does have an effect on safety feeling. Specifically, our results suggest that when a robot uses the JMM during handover, the safety feeling of the user increases. To distinguish whether this effect applies to both robot types, paired-samples t-tests were performed for each robot type. It was found that there was no significant difference in the feeling of safety between the two models JMM ($M=4.4, SD=0.51$) and LJST($M=4.27, SD=0.7$) for the humanoid robot; $t(14)=0.8, p=0.43$.  In contrast, the t-test of industrial robots showed a significant difference in the scores for JMM($M=3.8, SD=1.08$) and LJST($M=3.33, SD=0.97$); $t(14)=3.5$, $p=0.004$. These results suggest that the increase in the feeling of safety is more pronounced in industrial robots than in humanoid robots. However, the non-significant increase can also be attributed to the high average value of the models in the humanoid robot.
In order to test whether there was a different feeling of safety with the two robot models independent of the movement model, another paired-samples t-test was carried out. The results for the humanoid robot($M=4.33, SD=0.60$) and the industrial robot ($M=3.56, SD=1.04$) show a significant difference; $t(29)=4.892$, $p=0.000$.

\section{CONCLUSION}
The aim of this work was to determine a joint model for robots that makes a handover look more human-like. In addition, it was investigated whether humans perceive differences in robot movements. For the model, the movements of joints during human transfers were first observed and examined. Subsequently, the observations were approximated to characteristic functions using curve-fitting. These functions were used to transfer the movements to the robots. Finally, a study was carried out in which the movements of test persons were evaluated.
The results of the study showed for humanoid robots that most users only noticed a difference between the movements when they are pointed out. If the difference is recognized, the movements of the JMM were perceived as more human-like. The sequence and a partial gesture, which is similar a prompting gesture, were mentioned as a characteristic of the human likeness. To validate the results, some of the participants were tested as an control group. An additional evaluation criterion was the test persons' sense of safety. The JMM was found to be significantly safer if the models were viewed independently of the robot. Considering the robot type, a significant difference was only found in the industrial robot, whereas no significant difference can be detected in the humanoid robot.
In future work, we want to combine the joint model with a trajectory model in order to achieve an exact position control. The resulting model has to be compared with existing human-like solutions to determine performance. Possible subconscious signals of the subjects are also to be recorded in order to obtain results for the feeling of security. Another limitation of this study was the positioning of the handover pose. This was static for all persons. A dynamic approach is planned to adapt the poses to people.


\bibliography{sample}   
\bibliographystyle{IEEEtran}

\addtolength{\textheight}{-12cm}   

\end{document}